# An Efficient and Explainable Transformer-Based Few-Shot Learning for Modeling Electricity Consumption Profiles Across Thousands of Domains.

**Weijie Xia[1], Gao Peng[2], Chenguang Wang[3], Peter Palensky[1], Eric Pauwels[2], Pedro P. Vergara[1]** [*]

[1] Intelligent Electrical Power Grids (IEPG), Delft University of Technology, Delft, The Netherlands
[2] Centrum Wiskunde & Informatica (CWI), Amsterdam, The Netherlands
[3] Alliander N.V., Arnhem, The Netherlands

{W.Xia, P.P.VergaraBarrios, P.Palensky}@tudelft.nl, {Gao.Peng, Eric.Pauwels}@cwi.nl, chenguang.wang@alliander.com

## Abstract

Electricity Consumption Profiles (ECPs) are crucial for operating and planning power distribution systems, especially with the increasing numbers of various low-carbon technologies such as solar panels and electric vehicles. Traditional ECP modeling methods typically assume the availability of sufficient ECP data. However, in practice, the accessibility of ECP data is limited due to privacy issues or the absence of metering devices. Few-shot learning (FSL) has emerged as a promising solution for ECP modeling in data-scarce scenarios. Nevertheless, standard FSL methods, such as those used for images, are unsuitable for ECP modeling because (1) these methods usually assume several source domains with sufficient data and several target domains. However, in the context of ECP modeling, there may be thousands of source domains with a moderate amount of data and thousands of target domains. (2) Standard FSL methods usually involve cumbersome knowledge transfer mechanisms, such as pre-training and fine-tuning, whereas ECP modeling requires more lightweight methods. (3) Deep learning models often lack explainability, hindering their application in industry. This paper proposes a novel FSL method that exploits Transformers and Gaussian Mixture Models (GMMs) for ECP modeling to address the above-described issues. Results show that our method can accurately restore the complex ECP distribution with a minimal amount of ECP data (e.g., only 1.6% of the complete domain dataset) while it outperforms state-of-the-art time series modeling methods, maintaining the advantages of being both lightweight and interpretable. The project is open-sourced at https://github.com/xiaweijie1996/TransformerEM-GMM.git.

## 1 Introduction

Electricity Consumption Profiles (ECPs) refer to the daily (or other specified periods) time series data of electricity usage, reflecting the volatility of human consumption behavior. ECP modeling involves understanding and modeling the complex distribution of ECP data. This modeling has significant applications in the energy sector. For instance, the modeled distribution of ECP for households or areas can be used to generate additional ECP data, aiding in model training for electricity consumption prediction (Zhou et al.

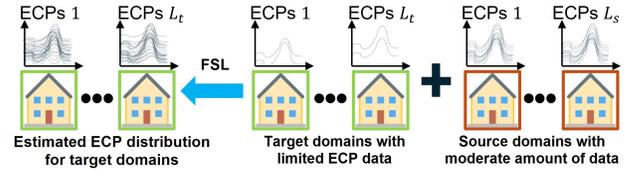

Figure 1: Modeling ECP distribution across many domains (households) using FSL. $L_s$ and $L_t$ are the numbers of source and target domains (households), respectively.

2020) and load monitoring (Francou et al. 2023). Understanding ECP distribution is also valuable for anomaly detection (Wang et al. 2020), risk analysis (Duque et al. 2023), energy supply and demand management (Shao, Pipattanasomporn, and Rahman 2012), and energy system control (Zhang et al. 2019; Hou et al. 2024). With the prevalence of deep learning (DL), models such as generative adversarial networks (GANs), variational autoencoders (VAEs), diffusion, and flow-based models are adopted for ECP distribution modeling (Xia et al. 2024b,a; Lin, Palensky, and Vergara 2024). However, these modeling approaches are relatively 'data-hungry' and usually assume sufficient ECP training data exists in the target domains. In practice, access to ECP data in the target domain can be limited due to various reasons such as privacy issues or absences of metering devices (McDaniel and McLaughlin 2009; Fang et al. 2023). In this paper, a domain refers to the ECP data collected from a terminal metering device in a residential household or building that has a unique electricity consumption pattern due to human behavior differences.

Few-shot learning (FSL) has emerged as a promising solution for ECP modeling in data-scarce scenarios. As demonstrated in (Yang, Liu, and Xu 2021), even with a limited number of samples, it is possible to calibrate distributions effectively for classification tasks. FSL has also been widely applied in images and audio generation (Kong et al. 2024; Ojha et al. 2021). This enlightens us to consider applying FSL in ECP modeling within data-scarce scenarios. Nevertheless, unlike a classic FSL task in image generation, in which there are usually several source domains with sufficient data and several target domains, ECP modeling often involves thousands of source domains (house-

---

holds) with moderate amounts of data and thousands of target domains (households). Furthermore, DL-based FSL typically requires model fine-tuning, which can be difficult to do effectively across a wide range of domains in tasks like ECP modeling. These issues let us ask *how can we develop a lightweight FSL method for ECP modeling?*.

Gaussian Mixture Models (GMMs) are widely applied across various distribution modeling tasks, including ECP modeling (Xia et al. 2024a). The most common way to estimate the parameters of GMMs is the EM algorithm. The advantages of GMMs include 1) they are lighter in computational complexity compared to DL models, 2) it is a whitebox model, and 3) similar to DL models, GMMs can theoretically approximate any distribution by increasing the number of components. Despite GMMs's advantages, GMMs as classical models seem isolated from FSL tasks, combining the advantages of FSL and GMMs remains an open problem. Recent work (Bertinetto et al. 2016; Ma et al. 2020) has shown that it is possible to train a learner where one DL model is used to predict the parameters of another DL model. This inspires us to question *whether a DL model can be used to assist in the parameter estimation of GMMs with limited samples as inputs*? Fig 1 demonstrates this idea. However, before addressing the above-mentioned question, we must first answer a preliminary question: *Which DL architecture is best suited for parameter estimation in GMMs*? Given the advantages of Transformers, such as their ability to capture dependencies and effectively knowledge transfer (Vaswani et al. 2017; Devlin et al. 2018), we select the Transformer as our foundational DL architecture for GMM parameter estimation. More details of the selection and adjustment of the Transformer are provided in Section 4.2.

Inspired by the questions and insights described above, we propose an FSL method for ECP modeling. First, we use a Transformer encoder architecture to acquire general knowledge from source domains. Then, we leverage this encoder to assist in the parameter estimation of GMMs in the target domains. We interpret the knowledge learned from the source and target domains as shifts in the mean and variance of the Gaussian components in GMMs, enhancing the interpretability of the obtained model. To the best of our limited knowledge, this is the first research to propose the FSL method for time series distribution modeling and, more specifically, for ECP distribution modeling in data-scarce scenarios across thousands of source and target domains. In summary, the contributions of this paper include:

- Firstly, we propose a lightweight and efficient method for FSL tasks that leverages a Transformer encoder architecture and GMMs. This method does not require fine-tuning in the target domain, thereby making it suitable for applications in the energy sector and potentially extending to other Internet of Everything (IoE) applications that involve numerous source and target domains (terminal devices) (Song et al. 2023).

- Secondly, we introduce a novel explainable FSL method that interprets knowledge learning as shifts in the means and variances of the Gaussian components, thereby increasing the model's explainability (Dwivedi et al. 2023).

- Lastly, we provide a new dataset for FSL research. Unlike widely used electricity datasets that are aggregated-level consumption data which exhibits stable patterns (Wu et al. 2022; Goswami et al. 2024), the new dataset contains volatile electricity consumption data of the individual household.

## 2 Related Work

### 2.1 Electricity Consumption Profile Modeling

ECP modeling is an active topic of research in the energy field, as accurate ECP models are essential for tasks such as power distribution systems state estimation (Angioni et al. 2015) and estimating the penetration rates of electric vehicles or solar panels (Ricciardi et al. 2018). GMMs and Copula are widely used methods for ECP modeling (Duque et al. 2021; Gemine et al. 2016). Recent works (Xia et al. 2024a) have shown that the DL models excel at capturing the temporal correlations of ECP, which is crucial for planning the necessary future investment of flexible power distribution systems. With the prevalence of DL, many deep generative models have been applied in ECP modeling. For instance, (Lin, Palensky, and Vergara 2024) utilized a diffusion model for high-resolution, 1-minute level ECP modeling. Additionally, (Wang and Zhang 2024; Xia et al. 2024b; Pan et al. 2019) employed conditional generative models to generate ECP data under varying weather conditions and customer characteristics. However, as mentioned earlier, current ECP modeling methods typically assume sufficient training data is available. In this context, ECP modeling in data-scarce scenarios remains an underexplored research problem.

### 2.2 Few-Shot Learning

Conventional FSL focuses on learning a discriminative classifier for tasks such as classification and detection (Yang, Liu, and Xu 2021; Zhang et al. 2021; Ma and Zhang 2019). ECP modeling, which involves modeling the distribution of ECP data, is essentially an FSL generation task (Song et al. 2023), and it has been extensively studied for data applications based on images, audio, and text (Kong et al. 2024; Zhao et al. 2022b; Ma et al. 2020; Chen et al. 2021). A typical FSL method usually involves pre-training in the source domain and fine-tuning in the target domain (Wang et al. 2018). This procedure can easily lead to overfitting in the target domain during fine-tuning. To address this, (Li et al. 2020) proposed an elastic weight consolidation in the loss function to prevent overfitting. Similarly, (Ojha et al. 2021) introduced a cross-domain correspondence mechanism to improve the diversity of model outputs and reduce overfitting. (Zhao et al. 2022a) proposed a method to adaptively preserve the knowledge learned in the source domain, considering the target domain. Despite the prevalence of FSL data applications based on images, audio, and other fields, its application to ECP modeling remains unexplored. Additionally, pre-training and fine-tuning are cumbersome for the massive number of domains expected in the energy sector. Therefore, a more efficient FSL method for ECP modeling is required.

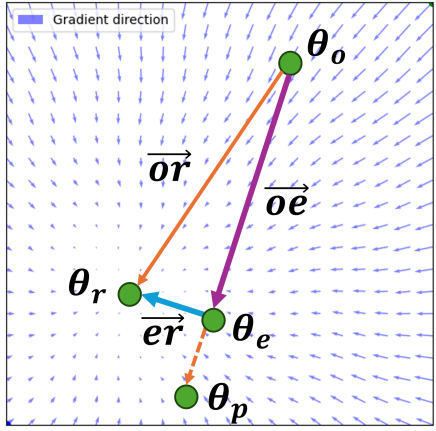

Figure 2: Our proposed method begins with $\theta_o$ as the initial parameter of the GMMs. Let $\theta_r$ be the optimal parameter for the target domain (or the estimated parameter assuming a complete ECP dataset). After applying the $z$-step EM algorithm on limited target-domain data, we obtain the estimated parameters $\theta_e$. If the GMMs converge on this limited data, we achieve $\theta_p$. Our approach uses a Transformer to predict $\vec{er}$ such that $\theta_r = \vec{er} + \theta_e$.

### 2.3 Time Series Modeling and Imputation

Time series data are ubiquitous across various industries. A fundamental task is to model the time series data distribution using a simplified statistical model, which facilitates tasks like imputation and forecasting (Chatfield 2000). Although many DL-based time series modeling methods have been developed in recent years, they often lack fo explanability (Lim and Zohren 2021). Due to security and safety reasons in power distribution systems applications, more explainable models are preferred when it comes to ECP modeling (Xu et al. 2023; Zhang, Cheng, and Yu 2024).

Time series imputation addresses the common issue of incomplete data in collected time series (Fang et al. 2023). Numerous methods have been proposed in this field (Bandara, Hyndman, and Bergmeir 2021; Wu et al. 2022; Goswami et al. 2024; Rasul et al. 2021). Imputation typically requires more data in the target domain than FSL tasks, for example, using 50% of the complete data to impute the remaining 50% (Goswami et al. 2024; Wu et al. 2022; Liu et al. 2023).

The task in this paper is similar to time series probabilistic imputation, where we treat each day of electricity consumption data as an individual sample and estimate its distribution using limited samples. Our proposed methods can also be extended to probabilistic imputation, as knowing the distribution and having partial observations turns it into a conditional generation problem (Tashiro et al. 2021). Due to the challenges standard FSL methods face in our task and its similarity to time series imputation, we use the method from (Wu et al. 2022) as a baseline for comparison.

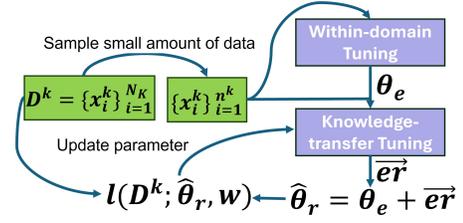

Figure 3: Training/inference process of one domain. $\hat{\theta}_r = \{\mu_j, \sigma_j\}_{j=1}^{J}$ represents the predicted parameter of GMMs, $\mathbf{w}$ is the weights of components, $l(\cdot)$ is the loss function. In the training process, $\mathcal{D}^k \in \mathcal{S}$. In the inference process, $\mathcal{D}^k \in \mathcal{T}$, and only $\hat{\theta}_r$ is predicted without loss computation and parameter updating for the Transformer.

## 3 Problem Formulation
### 3.1 Preliminaries

In ECP modeling, a typical daily ECP sample consists of $T$ discrete time steps. For example, ECP data with a resolution of 60 minutes is characterized by a $T = 24$ time step (one for each hour), making one ECP sample a 24-dimensional point. Each dimension corresponds to a specific value of active power consumption of a time step. In general, the ECP dataset of $k$-th domain (household) can be described as a

$$\mathcal{D}^k = \{\mathbf{x_i}^k\}_{i=1}^{N^k} = \{(x_{1,i}^k, ..., x_{T,i}^k)\}_{i=1}^{N^k}, \quad \mathbf{x_i}^k \in \mathbb{R}^{1 \times T}, \tag{1}$$

where $x_{t,i}^k$ is the active power consumption of $i$-th ECP sample (day) and $k$-th domain at $t$-th time step, $\mathbf{x_i}^k = (x_{1,i}^k, ..., x_{T,i}^k)$ is the $i$-th sample in $k$-th domain, $N^k$ is the number of samples (days) of $k$-th domain. Therefore, the ECP dataset from all domains can be expressed as

$$\mathcal{D} = \{\mathcal{D}^k\}_{k=1}^{K}, \tag{2}$$

where $K$ is the amount of the domains (households).

### 3.2 Few-shot Learning Problem Formulation

Assuming we have source domain collection $\mathcal{S} \subset \mathcal{D}$ and target domain collection $\mathcal{T} \subset \mathcal{D}$, where $\mathcal{S} \cap \mathcal{T} = \emptyset$. We aim to train a Transformer $f_{\theta 1}$ on $\mathcal{S}$, where we have access to a moderate amount of ECP data, for example, $250 \leq N^{k_s}$ for any $k_s$-th domain in $\mathcal{S}$. Our goal is to generalize $f_{\theta 1}$ to $\mathcal{T}$ to predict the parameters of GMMs with limited access to the data. For example, we sample $n^{k_t}$ ($0 < n^{k_t} \leq 25$) ECP samples from $k_t$-th domain in $\mathcal{T}$ as input of $f_{\theta 1}$. We aim for the GMMs with predicted parameters to represent the $k_t$-th domain's real ECP distribution accurately.

## 4 Methodology

FSL methods usually include pre-training, which involves acquiring general knowledge, and fine-tuning, which involves acquiring domain-specific knowledge. Our method has similar processes but in reverse order. Fig 2 demonstrates the overall idea of our approach. Let $\theta_o$ be the initial parameter of the GMMs, and $\theta_r$ be the optimal parameter of GMMs for a target domain (or the estimated parameter of GMMs assuming having complete ECP dataset of this domain). Our objective is to find the vector $\vec{or}$ in the parameter space such that

$$\theta_r = \vec{or} + \theta_o. \quad (3)$$

we write $\theta$ instead of $\vec{\theta}$ for simplicity. However, directly using a Transformer to predict $\vec{or}$ can be difficult and unstable, as the distance $||\theta_o - \theta_r||$ in the parameter space of GMMs can be very large. To address this issue, we propose an alternative approach.

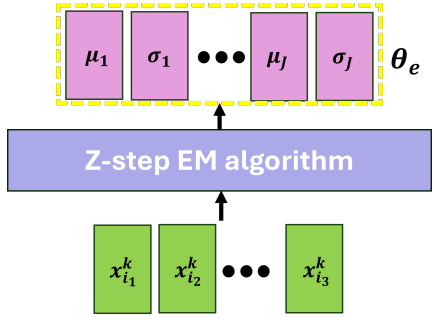

Figure 4: $z$ step EM algorithm (Within-domain Tuning) aims at learning target-domain specific knowledge. $x_i^k$ is the input ECP samples of $k$-th domain and $i$-th sample, in this process. $J$ is the number of components of GMMs, each $\mu$ and $\sigma$ are the parameters of a Gaussian component in GMMs.

First, we perform the $z$ step EM algorithm on the target domain data, which has a limited amount of data. Let $\theta_e$ be the estimated parameters of the GMM after $z$ step EM as shown in Fig 2. In this process, we do not aim for the GMMs to converge on the target domain data (reaching $\theta_p$), but instead, stop early to achieve an approximate minimal $||\vec{er}||$. This step can be considered as fine-tuning in the target domain, with early stopping to prevent overfitting. Second, we train a transformer encoder on the source domain to learn to predict $\vec{er}$. We can then compute $\vec{or}$ by

$$\vec{or} = \vec{oe} + \vec{er}. \quad (4)$$

In this process, $\vec{er}$ represents the transferred knowledge from the source domain considering $\theta_e$, and $\vec{oe}$ captures the target-domain specific knowledge. For convenience, we refer to the first process ($z$-step EM algorithm) as *Within-domain Tuning* and the second process as *Knowledge-transfer Tuning*.

Fig 3 summarizes how Within-domain Tuning and Knowledge-transfer Tuning function during both training and inference. During training, we randomly sample a batch of source domains from $\mathcal{S}$. Next, from these source domains, we randomly sample a small number (e.g., $4 \leq n^{k_s} \leq 25$) of ECP samples from each domain. These ECP samples are used in Within-domain Tuning to obtain the corresponding $\theta_e$ for each domain. Subsequently, the ECP samples and $\theta_e$ are fed into the encoder to predict the $\vec{er}$. The encoder is updated based on the loss described in Section 4.3. During inference, we follow the same procedure but use domain data from $\mathcal{T}$ to predict $\hat{\theta}_r$ without loss computation and parameter updating for the Transformer.

**Algorithm 1: Within-domain Tuning of One Domain**

**Require:** Sampled ECP data **x**, initial parameters $\theta_o$, fixed weights **w**, iteration steps $z$.
1: Initialize $\theta = \theta_o$
2: **for** $i = 1$ to $z$ **do**
3:    **E-step:**
4:    **for** $j = 1$ to $J$ **do**
5:       Compute the responsibility $\gamma_{ij}$:

$$\gamma_{ij} = \frac{w_j \mathcal{N}(x_i|\mu_j, \sigma_j)}{\sum_{k=1}^{J} w_k \mathcal{N}(x_i|\mu_k, \sigma_k)}$$

6:    **end for**
7:    **M-step:**
8:    **for** $j = 1$ to $J$ **do**
9:       Update the $\mu_j$:

$$\mu_j = \frac{\sum_{i=1}^{N} \gamma_{ij} x_i}{\sum_{i=1}^{N} \gamma_{ij}}$$

10:     Update the $\sigma_j$:

$$\sigma_j^2 = \frac{\sum_{i=1}^{N} \gamma_{ij}(x_i - \mu_j)^2}{\sum_{i=1}^{N} \gamma_{ij}}$$

11:    **end for**
12: **end for**

In the following sections, we provide a detailed explanation of the Within-domain Tuning and Knowledge-transfer Tuning processes.

### 4.1 Within-domain Tuning

This section demonstrates how $\theta_e$ is obtained by Within-domain Tuning. As mentioned before, Within-domain Tuning is essentially the $z$ step EM algorithm of GMM on a small number of ECP samples.

In our method, we set the weights **w** of GMMs components to be fixed during the EM iteration, following the simple rule

$$\mathbf{w} = \left[ \frac{1}{\sum_{j=1}^{J} j}, \frac{2}{\sum_{j=1}^{J} j}, \ldots, \frac{J}{\sum_{j=1}^{J} j} \right], \quad (5)$$

where $J$ represents the number of components of GMMs. In this context, fixing the **w** will not affect the expressiveness of the GMMs, as expressiveness remains consistent (or could be enhanced) as the number of components increases (McLachlan, Lee, and Rathnayake 2019). We fix **w** because we find that a Gaussian component's mean and covariance matrix are highly sensitive to its weights. Therefore, fixing the weights helps stabilize the entire learning process. For a similar reason, we apply the same initial parameters $\theta_o$ throughout the learning process.

We also set each Gaussian component to be spherical, meaning the covariance matrix can be expressed as $\text{diag}(\sigma)$, where $\sigma$ is a vector. This approach aligns the shapes of $x$, $\mu$ (mean vector), and $\sigma$, as $x_i^k$ and $\mu$ have the shape $1 \times T$.

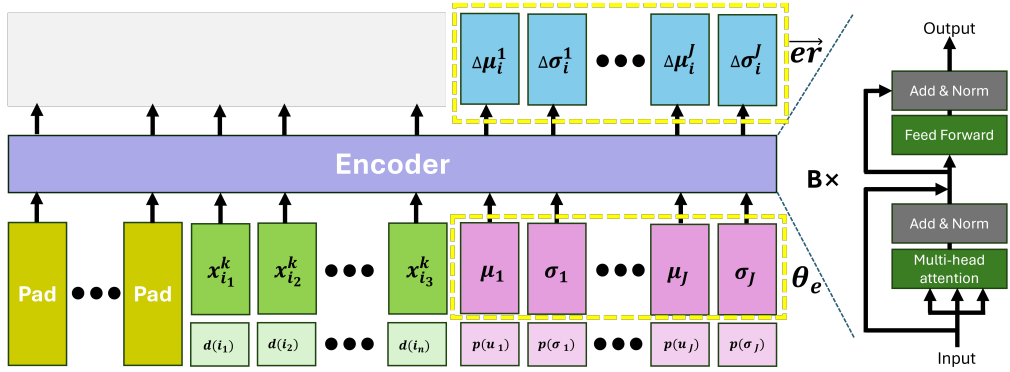

Figure 5: Knowledge-transfer Tuning process. Sampled ECP samples are fed into the encoder together with corresponding $\theta_e$ to predict $\vec{er}$. $d(\cdot)$ represents the date embedding, which indicates the day of the year for $x_i$, and $p(\cdot)$ represents the parameter information, indicating which Gaussian component of the GMMs $\mu$ and $\sigma$ belong to. We use the **[Pad]** token to align the shapes of inputs with different amounts $n^k$.

By setting the Gaussian components to be spherical, the covariance matrix can also be expressed as a $1 \times T$ vector $\sigma$, allowing us to treat $x$, $\mu$, and $\sigma$ as $1 \times T$ shaped tokens for the input and output of the Transformer as shown in Fig 4. Algorithm 1 shows how each component's $\mu$ and $\sigma$ are computed in Within-domain Tuning. Finally, We propose the following equation to determine $z$ in ECP modeling

$$z = \text{int}\left(e^{\beta n^k}\right), \qquad (6)$$

where $n^k$ is the number of ECP samples used in Within-domain Tuning, $\beta$ is a parameter set to $0.015$, which is the result of empirical testing in $\mathcal{S}$, $\text{int}(\cdot)$ means the integer part of a value.

### 4.2 Knowledge-transfer Tuning

This section demonstrates how to use a Transformer to predict $\vec{er}$. We adopt the encoder part of the Transformer for our model. Fig 5 illustrates how $\vec{er}$ is obtained in the Knowledge-transfer Tuning process.

We use a **[Pad]** token to align the shapes of inputs with varying sizes $n^k$. The encoder's parameters are updated based on the loss function explained in Section 4.3. $d(\cdot)$, shown in Fig 5, represents the date information, which indicates the day of the year for $x_i$, while $p(\cdot)$ represents the parameter information, indicating which Gaussian component of the GMMs $\mu$ and $\sigma$ belong to.

Regarding the design of the encoder, we use RMSNorm (Zhang and Sennrich 2019), and we do not use positional embedding. Considering a set of samples from the $k$-th domain and assuming an ***iid*** condition, the order of the samples does not affect the distribution and corresponding parameters of the GMMs. For instance, if $n^k = 20$, incorporating positional encoding would result in $20! > 2e + 18$ possible arrangements, significantly expanding the input space. Omitting positional encoding reduces these $20!$ cases to $1$ case, thus enabling a much more efficient learning process. This is the key reason that motivates us to use the Transformer encoder instead of other DL architectures.

### 4.3 Loss Design

Since GMMs are a white-box model, we can directly compute the log-likelihood based on the predicted parameters of the GMMs. The loss is defined as the negative log-likelihood. The loss function is given by

$$\log \mathcal{L}(\mathcal{D}^k \mid \hat{\theta}_r, \mathbf{w}) = -\sum_{i=1}^{N} \log \left( \sum_{j=1}^{J} w_j \mathcal{N}(\mathbf{x}_i^k \mid \mu_j, \sigma_j) \right), \qquad (7)$$

where $w_j \in \mathbf{w}$ is the fixed weight of the $j$-th Gaussian component, $N$ is the number of data points in the domain, $J$ is the number of Gaussian components, $\mathcal{N}(\mathbf{x}_i^k \mid \mu_j, \sigma_j)$ is the Gaussian probability density function for the $j$-th component with $\mu_j$ and $\sigma_j$, $\mathcal{D}^k$ is the complete domain dataset.

## 5 Experiments

### 5.1 Experiment Setting

**Data** The data used consists of hourly resolution electricity consumption data from individual households in the UK, Australia, Germany, and the Netherlands. The dataset includes approximately 20 thousand households in total. Due to varying household data lengths, we sample 250 days of ECP data from each household to create a domain. This ensures that each domain has an equal amount of ECP data. To fully utilize all available data, if a household has significantly more than 250 days of data, we sample multiple times, treating each sampled set as an individual domain. As a result, the dataset contains approximately 49 thousand domains. We carefully split the data into training, testing, and validation sets with a ratio of $0.8$, $0.1$, and $0.1$, ensuring that the augmented domain remains within the same set.

**Experiment Design** We find it challenging to identify an FSL method for images or audio that can be directly applied to our task. As mentioned in Section 2.3, given that time-series imputation is the closest task to our own, we aim to provide a clear comparison of the advantages of our proposed methods by selecting TimesNet (Wu et al. 2022), a popular model for time series imputation, as our baseline. Unlike standard data imputation tasks where the model has access to 80% to 50% of the complete time points, in this task, we provide both our model and TimesNet with only 1.6% to 10% of the complete time points during training. Our primary objective is to model the distribution of ECP data rather than to impute missing data. Therefore, we employ Maximum Mean Discrepancy (MMD) as the loss function to train TimesNet instead of commonly used Mean Squared Error (MSE) loss, eliminating the impact of the ECP sample order. We employed TimesNet to impute missing data, subsequently dividing the imputed time series into ECP samples to compare their distribution against the domain's complete dataset. Similarly, we generated ECP samples from predicted GMMs using our method to evaluate distribution differences relative to the domain's complete data.

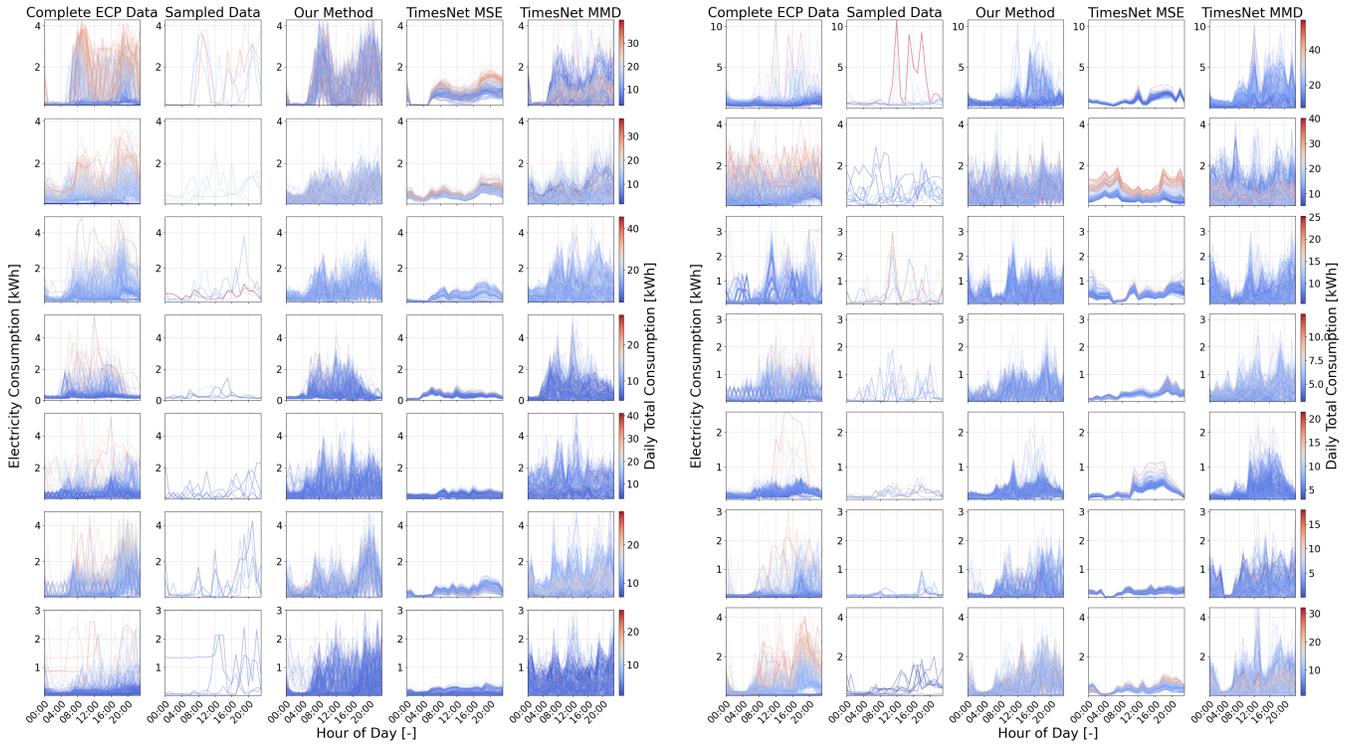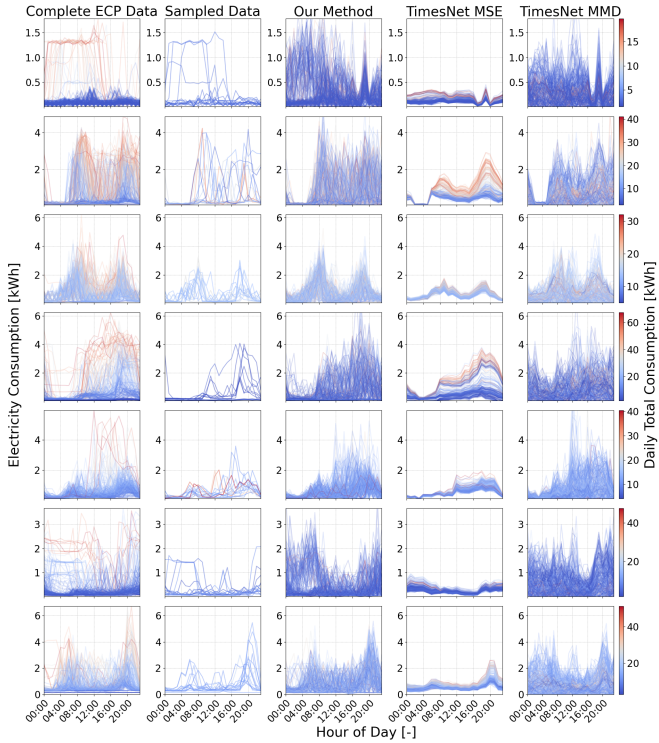

Figure 6: In each subfigure, every row represents the experimental results for a specific target domain. Each row, from left to right, includes (1) the complete ECP data of the domain, (2) the sampled ECP data used as input for our model and two TimesNets, (3) the ECP data generated from GMMs whose parameters are predicted by our method, (4) results generated by TimesNet using MSE loss, and (5) results generated by TimesNet using MMD loss. The color of the curves is related to the daily total electricity consumption.

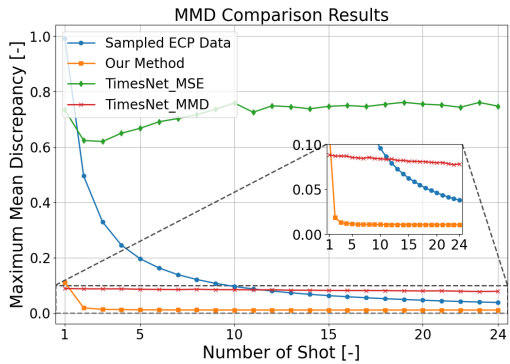

Figure 7: Comparison of MMD values. For each shot, we sample 100 target domains from $\mathcal{T}$ and compute the average MMD values of $n^{k_t}$ (number of shots) from 1 to 24.

**Evaluation Methods** Regarding the evaluation metrics, we use MMD, Kullback-Leibler (KL) divergence, Wasserstein distance (WD), MSE of the mean, and Kolmogorov-Smirnov (KS) distance to evaluate the distribution differences, following the methodologies outlined in (Toth, Bonnier, and Oberhauser 2020; Lin, Palensky, and Vergara 2024). A small value of these metrics indicates better performance.

**Hyperparameters and Training** We use one NVIDIA V100 GPU to train our model. For the training process, we utilize cyclical learning rates (Smith 2017), with the highest and lowest learning rates set to $1e-3$ and $1e-5$, respectively. The batch size is $128$. Both our model and TimesNet have a similar parameter scale, approximately 4.5 million parameters. The number of components of the GMMs, $J$, is set to 6.

## 5.2 Results

In this section, we present and discuss the experimental results of our proposed method. Fig 6 presents part of our experimental results of different $n^{k_t}$ (number of shots or number of sampled ECP samples as model input). Fig 6 shows that our method effectively captures human electricity consumption behavior volatility and restores the ECP distribution with minimal input samples. In contrast, TimesNet MSE struggles with volatile time series patterns, primarily generating ECP samples in high-likelihood areas. Although TimesNet MMD performs better than TimesNet MSE, it occasionally fails to restore the ECP distribution accurately. Furthermore, it is also noted that the predicted GMMs sometimes fail to capture the highest peaks and temporal correlations. Some peaks generated by GMMs are not as high as those in the original ECP data in Fig 6. Additionally, the results from GMMs tend to have lower daily total consumption, as indicated by the cooler colors of the curves. This finding aligns with (Xia et al. 2024a), which relates to the inherent limitation of GMMs. Table 1 provides a quantitative summary of the comparison results, demonstrating that our method typically achieves significantly smaller metric values compared to the alternatives. Fig 7 shows the comparison results of $n^{k_t}$ range from 1 to 24, where it can be seen that the estimated distribution from our method is closer to the original distribution, especially when the number of shots $1 < n^{k_t} \leq 14$, compared to the sampled ECP data and TimesNet MMD.

Table 1: Results of evaluation metrics in the validation set (average of 100 target domains).

| Method | MMD | KL | KS | WD | MSE.M |
|---|---|---|---|---|---|
| 4-shot | | | | | |
| Sampled ECP | 0.2434 | 0.9812 | 0.4244 | 0.4092 | 0.0392 |
| TimesNet MSE | 0.4938 | 13.419 | 0.4307 | 0.3544 | 0.0716 |
| TimesNet MMD | 0.0868 | 1.527 | 0.1912 | 0.3299 | 0.0337 |
| *Our Method* | 0.0222 | 0.4177 | 0.1499 | 0.1530 | 0.0171 |
| 8-shot | | | | | |
| Sampled ECP | 0.1196 | 0.7277 | 0.3044 | 0.2999 | 0.0192 |
| TimesNet MSE | 0.6008 | 13.613 | 0.4241 | 0.3639 | 0.0316 |
| TimesNet MMD | 0.0854 | 1.4935 | 0.1888 | 0.3267 | 0.0332 |
| *Our Method* | 0.0154 | 0.3535 | 0.1370 | 0.1339 | 0.0099 |
| 16-shot | | | | | |
| Sampled ECP | 0.0579 | 0.4885 | 0.2353 | 0.2084 | 0.0096 |
| TimesNet MSE | 0.6656 | 14.579 | 0.4366 | 0.3817 | 0.0255 |
| TimesNet MMD | 0.0809 | 1.4251 | 0.1893 | 0.3228 | 0.0315 |
| *Our Method* | 0.0126 | 0.3496 | 0.1315 | 0.1272 | 0.0062 |
| 24-shot | | | | | |
| Sampled ECP | 0.0371 | 0.3935 | 0.1985 | 0.1666 | 0.0058 |
| TimesNet MSE | 0.6960 | 15.332 | 0.4496 | 0.3942 | 0.0246 |
| TimesNet MMD | 0.0780 | 1.3695 | 0.1876 | 0.3194 | 0.0303 |
| *Our Method* | 0.0116 | 0.3358 | 0.1253 | 0.1217 | 0.0046 |

## 6 Conclusion

In this paper, we present an FSL method for ECP modeling. Our experiments demonstrate that our method can accurately estimate the original ECP distribution using only, for example, four ECP samples, which constitute just 1.6% of the complete domain dataset. The other primary advantages of our method, which also underscore the contributions of this paper, are as follows 1) Our proposed method is lightweight and efficient compared to standard FSL methods as we do not have fine-tuning, making it highly applicable in scenarios involving thousands or even millions of domains, 2) Our model is highly explainable, as we interpret knowledge learning as the mean and variance shifts of

GMMs, which serve as the final model deployed in applications. In the future, we aim to enhance this learn-a-white-box method to include a more advanced model, such as Copula, to address the inherent limitations of GMMs.

# 7 Acknowledgement

This publication is part of the project ALIGN4Energy (with project number NWA.1389.20.251) of the research program NWA ORC 2020 which is (partly) financed by the Dutch Research Council (NWO).